\documentclass[conference]{IEEEtran}
\IEEEoverridecommandlockouts
\usepackage{cite}
\usepackage{amsmath,amssymb,amsfonts}
\usepackage{graphicx}
\usepackage{textcomp}
\usepackage{xcolor}
\usepackage{myMacros}
\usepackage{booktabs}
\usepackage{balance}
\usepackage{algorithm}
\usepackage{algorithmicx}
\usepackage{subcaption}
\usepackage{multicol}
\usepackage{amsbsy}
\usepackage{listings}
\usepackage{xparse}
\usepackage{cite}
\usepackage{booktabs}
\usepackage{algorithm}
\usepackage[noend]{algpseudocode}
\usepackage[export]{adjustbox}
\def\BibTeX{{\rm B\kern-.05em{\sc i\kern-.025em b}\kern-.08em
    T\kern-.1667em\lower.7ex\hbox{E}\kern-.125emX}}
    
\renewcommand{\baselinestretch}{.966}

\begin{document}

\title{Adversarial training with informed data selection
\thanks{This study was financed in part by the Coordena\c{c}\~{a}o de Aperfei\c{c}oamento de Pessoal de N\'{i}vel Superior - Brasil (CAPES) - Finance Code 001. This
work was also supported by the Swiss Government Excellence Scholarships for Foreign Students.}
}

\author{\IEEEauthorblockN{ Marcele O. K. Mendon\c{c}a$^\ast$, Javier Maroto$^\star$, Pascal Frossard$^\star$ and Paulo S. R. Diniz$^\ast$}
\IEEEauthorblockA{$\ast$ \textit{SMT - Signals, Multimedia, and Telecommunications Lab.} \\
{Universidade Federal do Rio de Janeiro, DEL/Poli \& PEE/COPPE/UFRJ}\\
P.O. Box 68504, Rio de Janeiro, RJ, 21941-972, Brazil, \\ $\star$ École
Polytechnique Fédérale de Lausanne (EPFL), Switzerland. \\
emails: \{marcele.kuhfuss,diniz\}@smt.ufrj.br, \{javier.marotomorales,pascal.frossard\}@epfl.ch}
}

\maketitle

\begin{abstract}

With the increasing amount of available data and
advances in computing capabilities, deep neural networks (DNNs)
have been successfully employed to solve challenging tasks in
various areas, including healthcare, climate, and finance.
Nevertheless, state-of-the-art DNNs are
susceptible to quasi-imperceptible perturbed versions of the
original images --
adversarial examples. These perturbations of the network input can lead to  disastrous implications in critical areas where wrong decisions can directly affect human lives. Adversarial training is the most efficient solution to defend the network against these malicious attacks. However, adversarial trained networks generally come with lower clean accuracy and higher computational complexity. This work proposes a data
selection (DS) strategy to be applied in the mini-batch training. Based on the cross-entropy loss, the most relevant samples in the batch are selected to update the model parameters in the backpropagation. The simulation results show that a good compromise can be
obtained regarding robustness and standard accuracy, whereas the
computational complexity of the backpropagation pass is reduced.

\end{abstract}


\section{Introduction}
\label{sec:intro}

Over the past decade, the amount of
available digital data has exponentially increased. Thanks to the advances in computing capabilities, deep neural networks (DNNs) have been successfully employed to solve challenging  image and natural language processing tasks.
However,
state-of-the-art DNNs are known to be highly vulnerable to
adversarial examples \cite{szegedy2013intriguing, goodfellow2014explaining}. These small but malicious perturbations of the network input can manipulate the trained model to produce incorrect predictions with high confidence, and some perturbations can even fool different network models \cite{moosavi2017universal}. Since adversarial attacks might lead to  disastrous implications in critical areas like healthcare \cite{esteva2019guide}, climate \cite{liu2016application}
and finance \cite{dixon2017classification}, defending against them is critical.

So far, adversarial training is the most effective approach to mitigate the effect of strong attacks like the Projected Gradient Descent (PGD) attack \cite{madry2017towards}, DeepFool \cite{moosavi2016deepfool}, and AutoAttack \cite{croce2020reliable}. Training the DNN with perturbed versions of the original
samples makes it possible to improve the accuracy on unseen
adversarial examples, also known as  \emph{robustness accuracy} \cite{ortiz2021optimism}. However, generating adversarial examples during training can be highly computationally intense since each sample is usually built with  several steps in the direction of the gradient as the model is trained. 
Moreover, adversarial training generally decreases the \emph{standard accuracy}, that is, the accuracy on clean samples \cite{zhang2019theoretically}. This \emph{robustness-accuracy} tradeoff is reported to be highly data-dependent, especially regarding the data distribution \cite{ding2019sensitivity} and its quality \cite{dong2021data}. Furthermore, we only have access to a training dataset which is not necessarily representative for the problem we aim to learn. In this case, we could avoid using the entire training data. Since the dataset is reduced, we can save several computations during backpropagation and speed-up training. 
This hypothesis was already investigated for standard training in \cite{ferreira2021data, mendoncca2021data}. In this work, we extend the work in \cite{ferreira2021data, mendoncca2021data} and apply it to the adversarial training case. From each mini-batch composed of both clean and adversarial samples, the proposed data selection algorithm selects the most relevant samples based on the cross-entropy loss. Since only the selected samples are used to update the model parameters in the backpropagation, the training time is reduced. The selection also balances the necessary amount of clean and adversarial samples required to yield satisfactory robustness and standard accuracy.

The paper is organized as follows. Section~\ref{sec:adv_train} presents
a brief overview of the adversarial training method and some notations.  In section~\ref{sec:proposed}, we propose a data selection technique for adversarial training. The proposed approach is tested via simulation results in section~\ref{sec:sim_res}. Finally, section~\ref{sec:conc} includes some conclusion remarks.

\section{Adversarial Training}\label{sec:adv_train}

Adversarial training continually creates and incorporates adversarial examples into the training process of a deep neural network classifier 
\begin{equation}
  f_{\boldsymbol{\theta}}(\mathbf{x}) : \Rset^{N} \rightarrow \{1 \cdots C\},
\end{equation}
with $\boldsymbol{\theta}$ weights, which maps an input image $\mathbf{x}$ to a label $y$ from a dataset 
\begin{equation}\label{eq:dataset}
 \mathcal{D} =    \{(\mathbf{x}(1), y(1)),(\mathbf{x}(2), y(2)),\cdots, (\mathbf{x}(M), y(M))\},
\end{equation}
with $C$ possible classes. Adversarial training attempts to solve the min-max optimization problem
  \begin{equation} \label{eq:opt_minmax}
 \begin{split}
  \text{min}_{\boldsymbol{\theta}} \frac{1}{\mathcal{|D|}} \sum_{\mathbf{x},y \in \mathcal{D}} \text{max}_{\boldsymbol{\eta}}~& \mathcal{L}(f_{\boldsymbol{\theta}}( \mathbf{x}+\boldsymbol{\eta}),y)\\
   \text{s.t}~&||\boldsymbol{\eta}||_p \leq \epsilon,
  \end{split}
 \end{equation}
where $\mathcal{L}(f_{\boldsymbol{\theta}}( \mathbf{x}+\boldsymbol{\eta}),y)$ is the loss function on the adversarial sample and $\boldsymbol{\eta}$ is a small perturbation constrained by $\epsilon$. 

Creating adversarial samples involves solving the inner maximization problem in equation (\ref{eq:opt_minmax}), in which the loss function $\mathcal{L}$ is maximized in an effort to change the prediction, that is,
$f_{\boldsymbol{\theta}}(\mathbf{x}+\boldsymbol{\eta}) \neq f_{\boldsymbol{\theta}}(\mathbf{x})$. The optimization constraints ensure that the distance between the adversarial and original example should be less than $\epsilon$ under a particular norm,  $||\boldsymbol{\eta}||_p \leq \epsilon$. The norms aim to quantify how imperceptible to humans an adversarial example is. Some examples of norms are the $l_0$ norm, $l_2$ norm, and $l_\infty$. 
We then briefly review the most popular methods to create adversarial examples.

Introduced by \cite{goodfellow2014explaining}, the Fast Gradient Sign Method (FGSM) attack generates adversarial examples by modifying the input towards the direction where the loss $\mathcal{L}$ increases
\begin{equation}
    \mathbf{x}' = \mathbf{x} + \epsilon \text{sign}(\nabla_{\mathbf{x}}\mathcal{L}(\boldsymbol{\theta}, \mathbf{x},y)),
\end{equation}
with $\text{sign}(\cdot)$ the sign function, and $\nabla_{\mathbf{x}}\mathcal{L}(\boldsymbol{\theta}, \mathbf{x},y)$ the loss gradient with respect to $\mathbf{x}$. One of the strongest $l_\infty$-bounded at-
tacks, the PGD attack \cite{madry2017towards} tries to solve the inner maximization problem in equation (\ref{eq:opt_minmax}) following an iterative procedure. At each step $i$, the adversarial example is updated as
\begin{equation}\label{eq:pgd}
    \mathbf{x}'_{i} = \text{clip}_{\mathbf{x} + \epsilon}(\mathbf{x}_{i-1} + \alpha \text{sign}(\nabla_{\mathbf{x}}\mathcal{L}(\boldsymbol{\theta}, \mathbf{x},y))),
\end{equation}
in which function $\text{clip}_{\mathbf{x} + \epsilon}(\cdot)$ clips the input at the positions around the predefined perturbation range. In the context of $l_2$-bounded attacks, Deepfool \cite{moosavi2016deepfool} is an iterative attack optimized for the $l_2$-norm  based on a
linear approximation of the classifier. Using geometry concepts, DeepFool
searches within the region of the space that describes the output of the classifier (polyhedron) for the minimal perturbation
that can change the classifiers decision. Among black-box attacks, one pixel attack  \cite{su2019one} is a $l_0$-bounded attack that employs differential evolution to create adversarial examples without knowing the network gradients and its parameters. Finally, the AutoAttack \cite{croce2020reliable} method consists of an ensemble
of four attacks: two versions of the PGD attack, the targeted version of the Fast Adaptive Boundary (FAB) attack \cite{croce2020minimally} and the black-box Square Attack \cite{andriushchenko2020square}.
Currently, AutoAttack and PGD attack are the most popular methods to test adversarial robustness. Since the PGD attack is less computationally intense than AutoAttack, we 
consider the PGD attack in this work. However, other attacks can be used with the proposed data selection.

With the inner maximization problem addressed, the  
outer minimization problem in equation (\ref{eq:opt_minmax}) is then solved to find the model parameters that minimize the loss on the generated adversarial examples. The original dataset $\mathcal{D}$ is split into small batches $\mathcal{B}$ and stochastic gradient descent (SGD) is employed to update the
model parameters 
 \begin{equation} \label{eq:sgd}
 \boldsymbol{\theta}_{t} = \boldsymbol{\theta}_{t-1} + \mu\frac{1}{\mathcal{|B|}} \sum_{\mathbf{x},y \in \mathcal{B}} \nabla_{\boldsymbol{\theta}}  \mathcal{L}(f_{\boldsymbol{\theta}}( \mathbf{x}+\boldsymbol{\eta}^*),y),
 \end{equation}
where the gradient is evaluated at the maximum point $\boldsymbol{\eta}^*$ found in the inner maximization problem, thanks to the Danskin's theorem \cite{danskin1966theory}.

\section{Proposed Data Selection for adversarial training}\label{sec:proposed}

When performing adversarial training, we are interested in 
learning a process or function $f(\cdot)$ that maps a data space $\mathcal{X}$ into an output space $\mathcal{Y}$. However, we do not have direct access to samples from $\mathcal{X}$ in order to
train the model according to the adversarial objective. We only have access to a subset $\mathcal{D}$ which is split into batches used to update the model parameters in equation~(\ref{eq:sgd}). However, there is no guarantee that this available subset or its batches consist of a good representation of the process $f(\cdot)$. In this regard, we propose a sampling strategy to select the most relevant samples to compose the batches in adversarial training.

\begin{figure}[!h]
    \centering
        \includegraphics[width=0.4\textwidth]{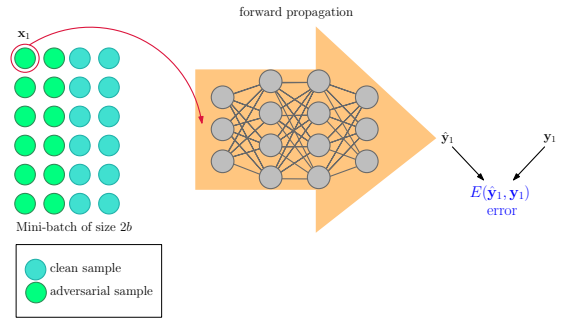}
      \caption{Forward propagation and error signal computation. }
            \label{fig:foward_error}
    \end{figure}

We first consider the entire original dataset $\mathcal{D}$ of input-output pairs in equation (\ref{eq:dataset}). Then,
at each mini-batch iteration, $b'$ clean samples are selected from the whole dataset to form the batch set $\mathcal{B'}$. By using PGD, $b'$ adversarial examples are generated from the samples in the set $\mathcal{B'}$ using equation~(\ref{eq:pgd}). The resulting mini-batch $\mathcal{B}$ is then composed of $b=2b'$ samples. The samples in the mini-batch flow through the network, the gradients are computed, and we obtain the network output as a one-hot-encoded vector $\mathbf{y}$, as shown in Figure~\ref{fig:foward_error}. In order to quantify the relevance of the samples in the mini-batch, we define the error signal
\begin{equation}\label{eq:loss_error}
E(\hat{\mathbf{y}}, \mathbf{y})= \sum_{c=1}^{C}e(\hat{y}_{c},y_{c}),
\end{equation}
which is based on the cross-entropy loss
\begin{equation}
 e(\hat{y}_{c},y_{c}) = \text{log} \left(\sum_{c=1}^C \text{exp}(\hat{y}_c) \right)- y_c,
\end{equation}
where $C$ is the number of classes. 

\begin{figure}[!h]
    \centering
        \includegraphics[width=0.4\textwidth]{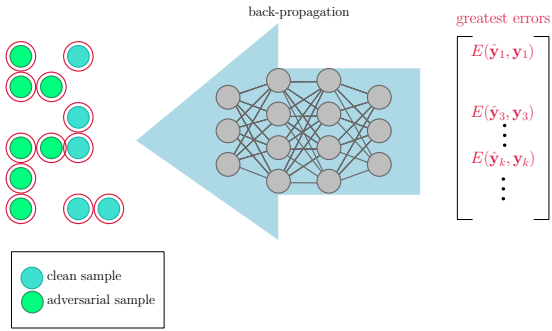}
      \caption{Selected samples being used in the backpropagation. }
            \label{fig:backprop}
    \end{figure}

As a rule, the closer to zero the error signal is, the less  informative  or  relevant  will be  the  contribution  of  the correspondent data pair to the parameter update in equation (\ref{eq:sgd}).  We then propose to select a portion $P_{\rm up}$ of the samples in $\mathcal{B}$ based on the higher error values in equation (\ref{eq:loss_error}), forming a selection set $\mathcal{S}$. After the forward propagation is completed, only the samples in $\mathcal{S}$ are used in the backpropagation to update the network parameters $\boldsymbol{\theta}$, as depicted in Figure~\ref{fig:backprop}. Since only a portion $P_{\rm up}$ of the samples are used to update the parameters, we can save some computations and we alleviate the training burden.

One question remains about how to choose an adequate $P_{\rm up}$ for our problem. As $P_{\rm up} \rightarrow 0 $, fewer samples are selected and we save more computations in the backpropagation. In this case, however, the selected samples might be insufficient lo learn the problem. For standard training, the most favorable $P_{\rm up}$ choice mainly depends on the dataset complexity \cite{ferreira2021data}. Simpler datasets like MINIST requires $P_{\rm up} = 0.3$, whereas for  more  complex  datasets  as CIFAR10, $P_{\rm up} = 0.5$ is a better choice. Thus, one option is to set a fixed $P_{\rm up}$ for the whole training process. In this way, we can set the amount of saved computations from the beginning.
Nevertheless, in cases where the dataset complexity is unknown and it is difficult to prescribe a $P_{\rm up}$ for all the epochs, an automatic $P_{\rm up}$ can be advantageous. In this way, we can obtain the $P_{\rm up}$ for each epoch in an adaptive manner as the training is performed. This can be achieved by considering the accuracy at each epoch as a criterion.
Hence, we can estimate the number of selected samples $P_{\rm up}$ at each epoch $t$.

\begin{equation}\label{eq:pup}
P_{\rm up}^{(t)} = (1 - \lambda_{acc}^{(t-1)}) P_{\rm up}^{(t-1)}
\end{equation}
where $P_{\rm up}^{(0)} = 1$ and  $\lambda_{acc}^{(t-1)}$ is the last available accuracy.
We need more samples in the mini-batch to improve learning when the accuracy is low, whereas fewer samples are required to continue the learning process when the accuracy increases.

As it will be shown in the simulations, updating the $P_{\rm up}$ using equation (9) accelerates the convergence for $P_{\rm up}^{(0)} = 1$ because, in this case, it selects more samples in the first epochs. Our motivation was to provide more samples to the model at the beginning to improve and accelerate its learning. Therefore, early stopping methods \cite{prechelt1998early} can be employed to further reduce the training time. Since we do not consider the early stopping approach in the simulations, we propose using a fixed prescribed $P_{\rm up}$ in this work. The main proposed algorithm is detailed in Algorithm~\ref{alg:ds}. 

\begin{algorithm}
\caption{Proposed Data Selection for adversarial training}
\label{alg:ds}
\begin{algorithmic}[1]

    \State Given dataset $\mathcal{D}$, mini-batch size $b'$, and prescribed $P_{\rm up}$
    \For{epoch = $1 \cdots T$}
    \For{mini-batch $\mathcal{B} \subset \mathcal{D}$}
    \State Create adversarial examples $\{x'_1, \cdots  x'_{b'} \}$ from clean samples $\{x_1, \cdots  x_{b'} \}$ using current state of the network and obtain 
     $\mathcal{B}' = \{x'_1, \cdots  x'_{b'}, x_1, \cdots  x_{b'} \}$;
    \State Forward propagation with samples in $\mathcal{B}'$;
    \State Compute the error signal for each sample in $\mathcal{B}'$ using equation~(\ref{eq:loss_error});
    \State Select the $P_{\rm up} \times 100 \%$ of the samples in $\mathcal{B}'$ with greatest error values;
    \State Update model parameters by back propagation using only the data samples in $\mathcal{S}$;
    \EndFor
    \EndFor

\end{algorithmic}
\end{algorithm}

\section{Simulation Results} \label{sec:sim_res}
In this section, we assess the performance of the proposed data selection method  in  the CIFAR10 dataset using the Resnet18 model. The PGD attack with $\epsilon = 8/255$, $\alpha = 0.01$ and 20 iterations is employed to build the adversarial examples. We consider the following methods in the simulations. The standard method trains only with clean samples with a mini-batch $\mathcal{B}$ of size $b = 256$. Also using $\mathcal{B}$ with $b = 256$, the robust method is trained only with adversarial examples. The DS robust method is trained with the selection set $\mathcal{S}$ of size $b = 256$, which is composed of both clean and adversarial samples, and it is obtained using our selection strategy with $P_{\rm up}$ fixed or varying. The random robust method is trained with a mini-batch of size $b = 256$, composed of clean and adversarial samples selected at
random. We also consider the selection method proposed in \cite{dong2021data} in which the samples are selected based on their learning stability. In this case, we used $50 \%$ of the samples with high quality in order to perform a fair comparison in terms of number of samples used.

\begin{figure}[!h]
    \centering
        \includegraphics[width=0.45\textwidth,trim={0 0 0 60},clip]{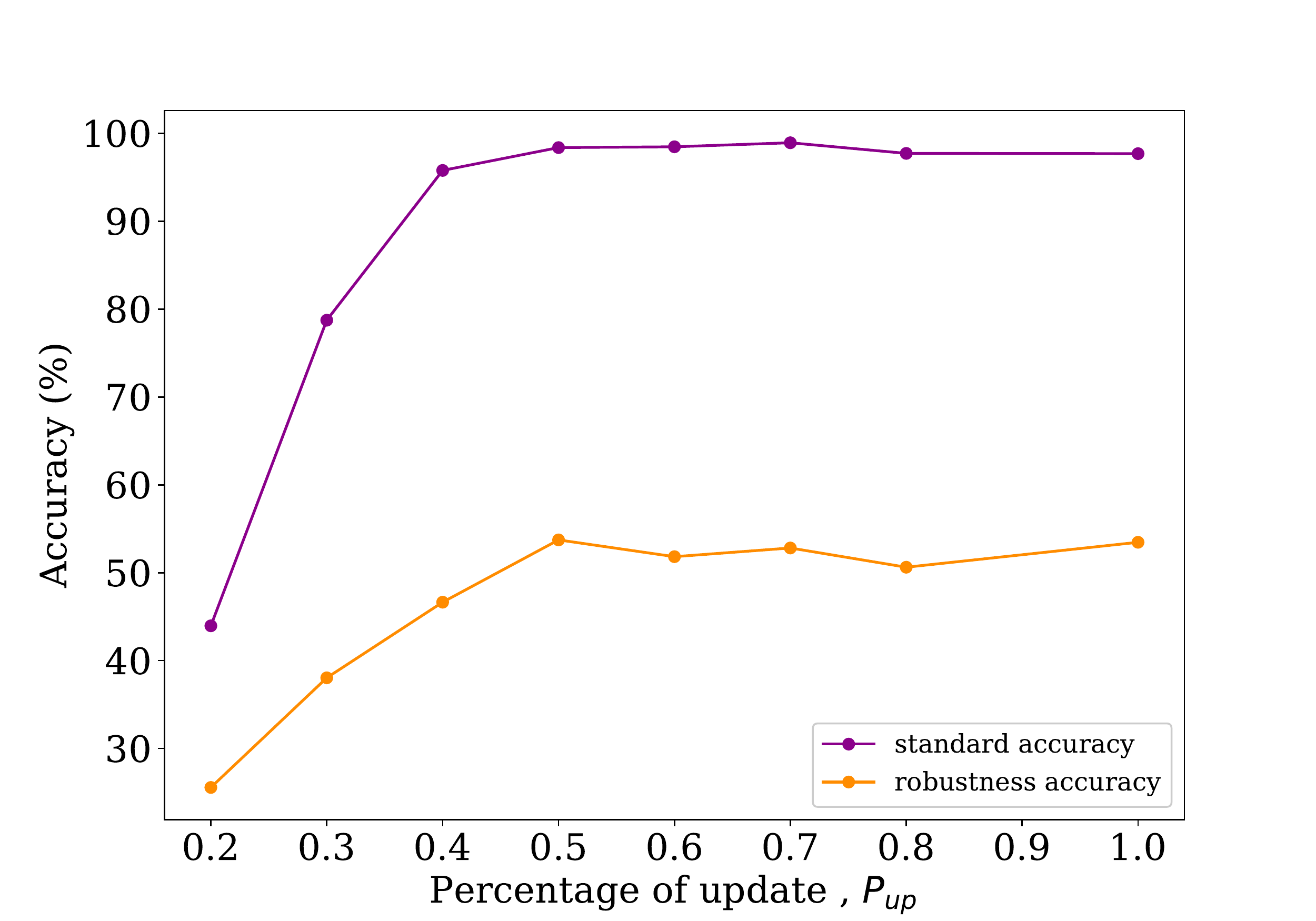}
      \caption{Evolution of the standard and robustness accuracy as the portion of selected samples $P_{\rm up}$ is varied.  }
            \label{fig:var_pup}
    \end{figure}

First, we vary the portion of selected samples $P_{\rm up}$ in Figure~\ref{fig:var_pup} to investigate the impact on the standard and robustness accuracy at the last epoch. By using $P_{\rm up} = 0.5$, we slightly outperform  the approach that consider all the samples ($P_{\rm up} = 1$) in terms of standard accuracy, with the benefit of requiring only $50 \%$ of the samples in the batch. In terms of robustness, the methods with $0.5 \leq P_{\rm up} < 1$  perform quite close to the method with  $P_{\rm up} = 1$. If we reduce $P_{\rm up} $ even further, we do not observe a gain in performance. In such case, the model would require more epochs to achieve the same performance or it would need more samples to learn the problem.

\begin{figure}[!h]
    \centering
        \includegraphics[width=0.45\textwidth,trim={0 0 0 60},clip]{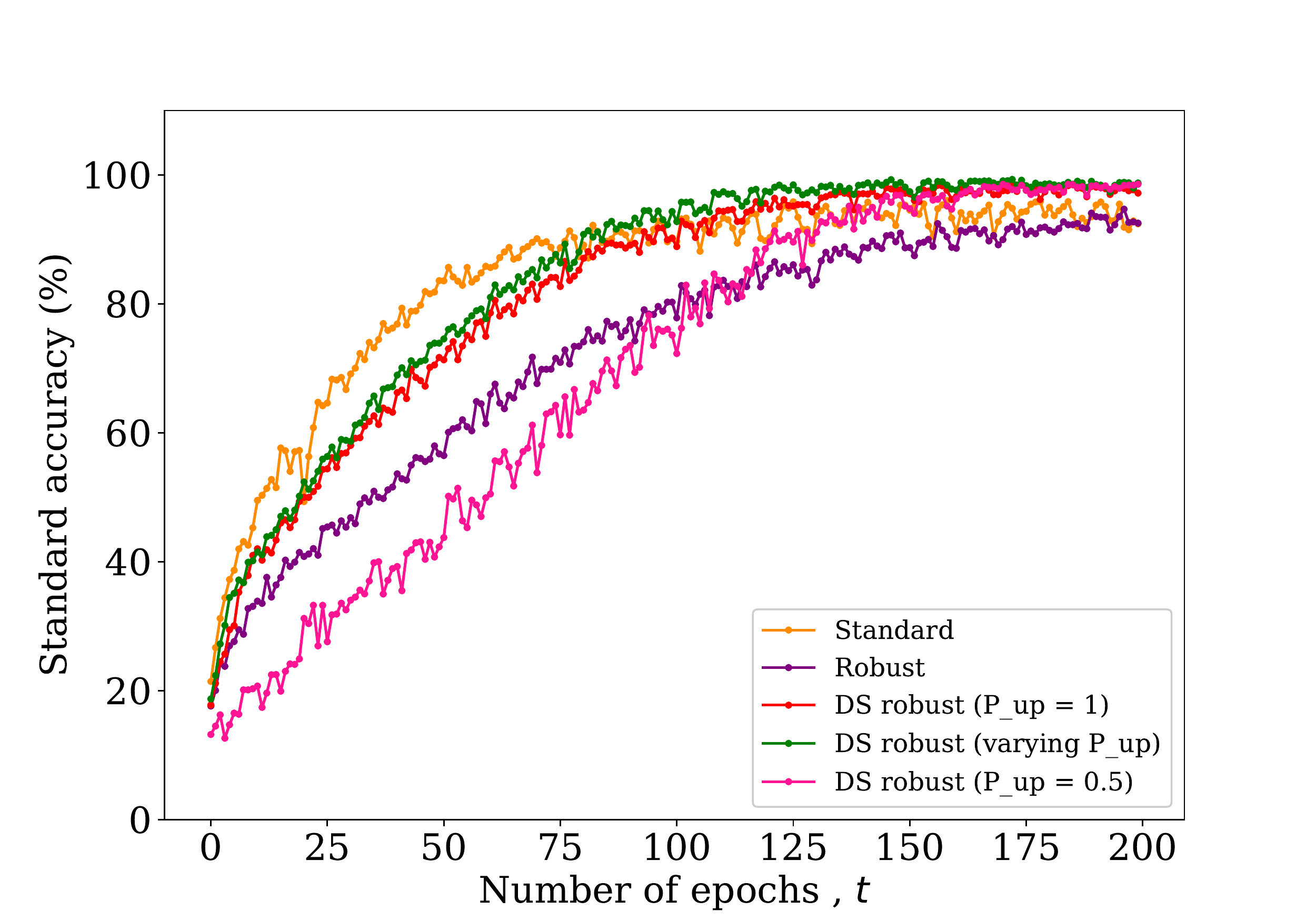}
      \caption{Standard accuracy as a function the number of epochs.  }
            \label{fig:acc_sta}
    \end{figure}

\begin{figure}[!h]
    \centering
        \includegraphics[width=0.45\textwidth,trim={0 0 0 60},clip]{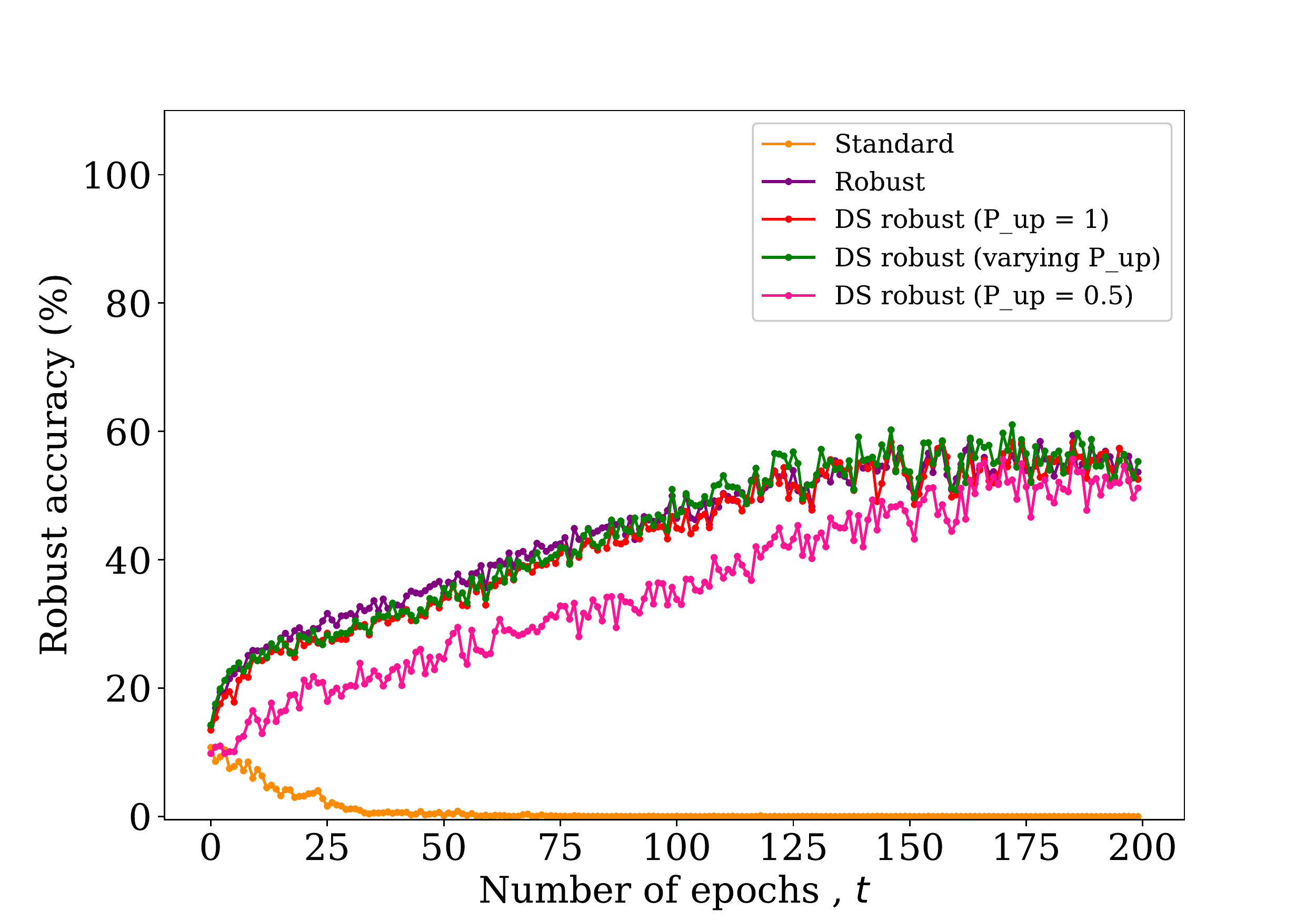}
      \caption{Robust accuracy as a function the number of epochs.  }
            \label{fig:acc_adv}
    \end{figure}

\begin{figure}[!h]
    \centering
        \includegraphics[width=0.45\textwidth,trim={0 0 0 20},clip]{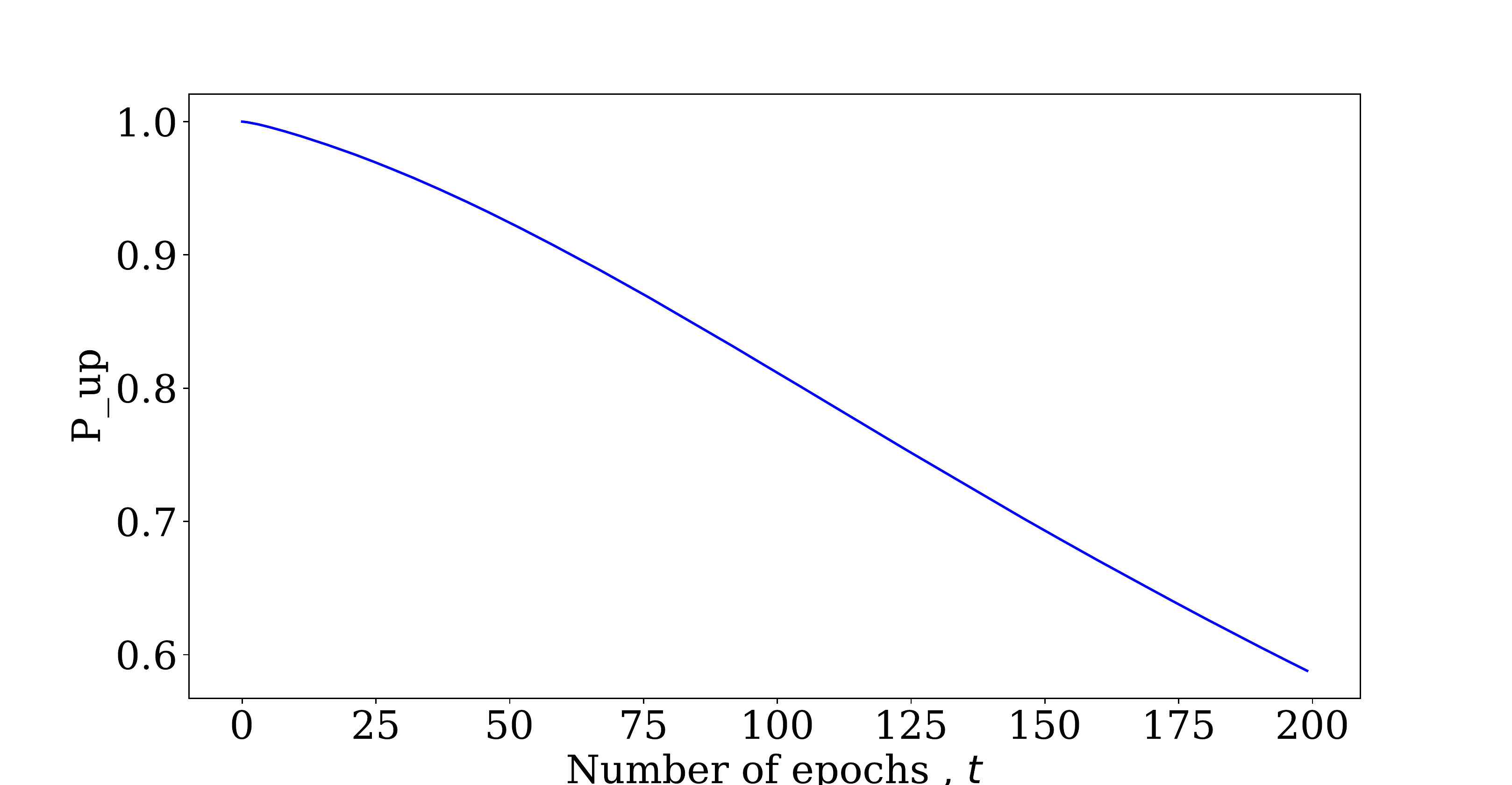}
      \caption{Portion of selected samples $P_{\rm up}$ obtained for each epoch following equation (\ref{eq:pup}). }
            \label{fig:pup}
    \end{figure}

We then evaluate the proposed DS robust method with varying $P_{\rm up}$ and compare it with the fixed $P_{\rm up} = 0.5$, the standard and robust methods in terms of standard and robust accuracy in
Figures~\ref{fig:acc_sta} and \ref{fig:acc_adv}. We show in Figure~\ref{fig:pup} the obtained $P_{\rm up}$ for each epoch following equation (\ref{eq:pup}). By using both a varying $P_{\rm up}$ and $P_{\rm up} =0.5$, we observe an improvement  in terms of standard accuracy when compared with the standard and robust methods. Moreover, reducing the number of samples in the mini-batch does not affect the robust accuracy, as shown in Figure~\ref{fig:acc_adv}.

\begin{figure}[!h]
    \centering
        \includegraphics[width=0.45\textwidth,trim={0 0 0 20},clip]{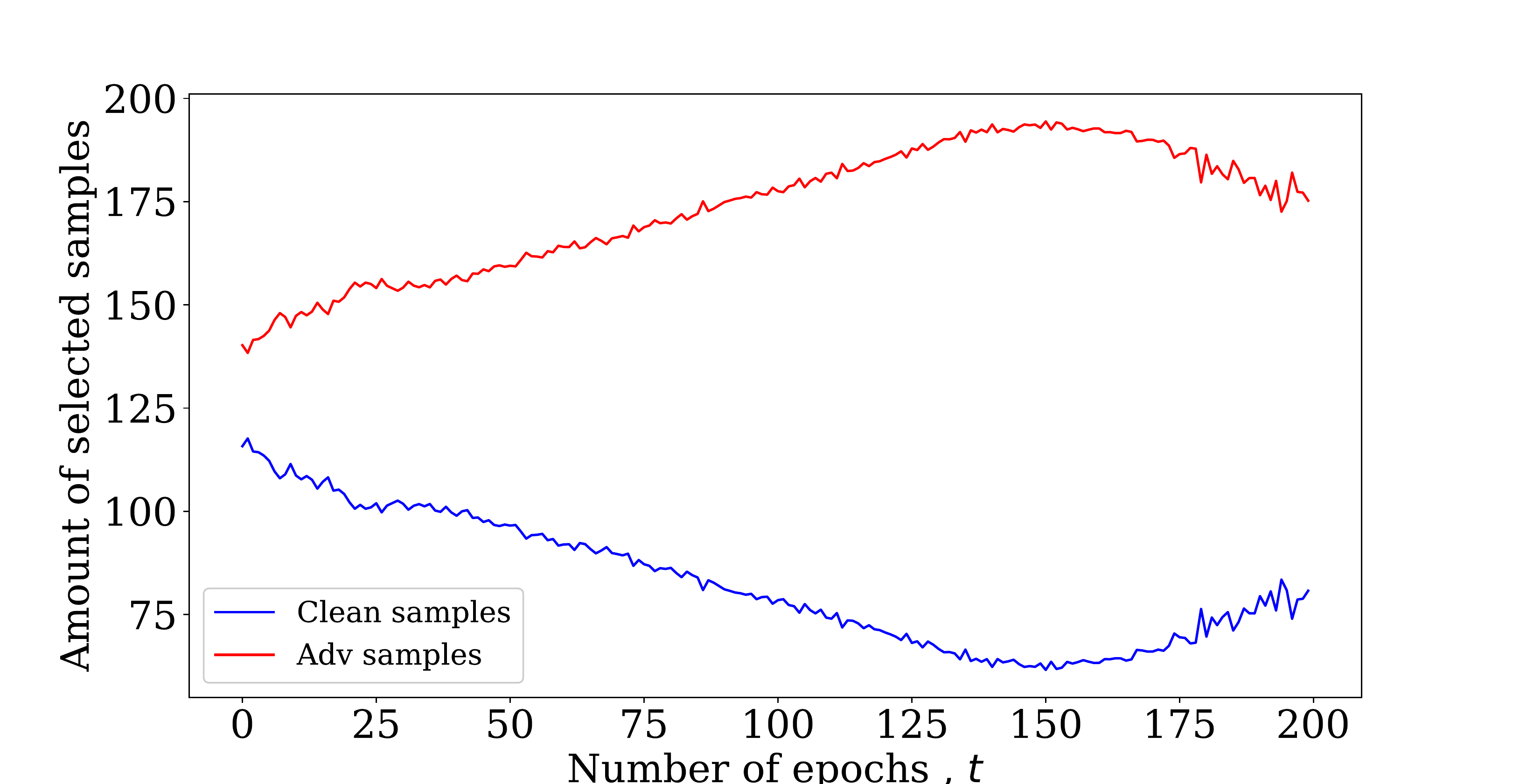}
      \caption{Averaged amount of selected clean and adversarial samples  at each epoch for $P_{\rm up} = 0.5$. }
            \label{fig:av}
    \end{figure}

\begin{figure}[!h]
    \centering
        \includegraphics[width=0.45\textwidth,trim={0 0 0 20},clip]{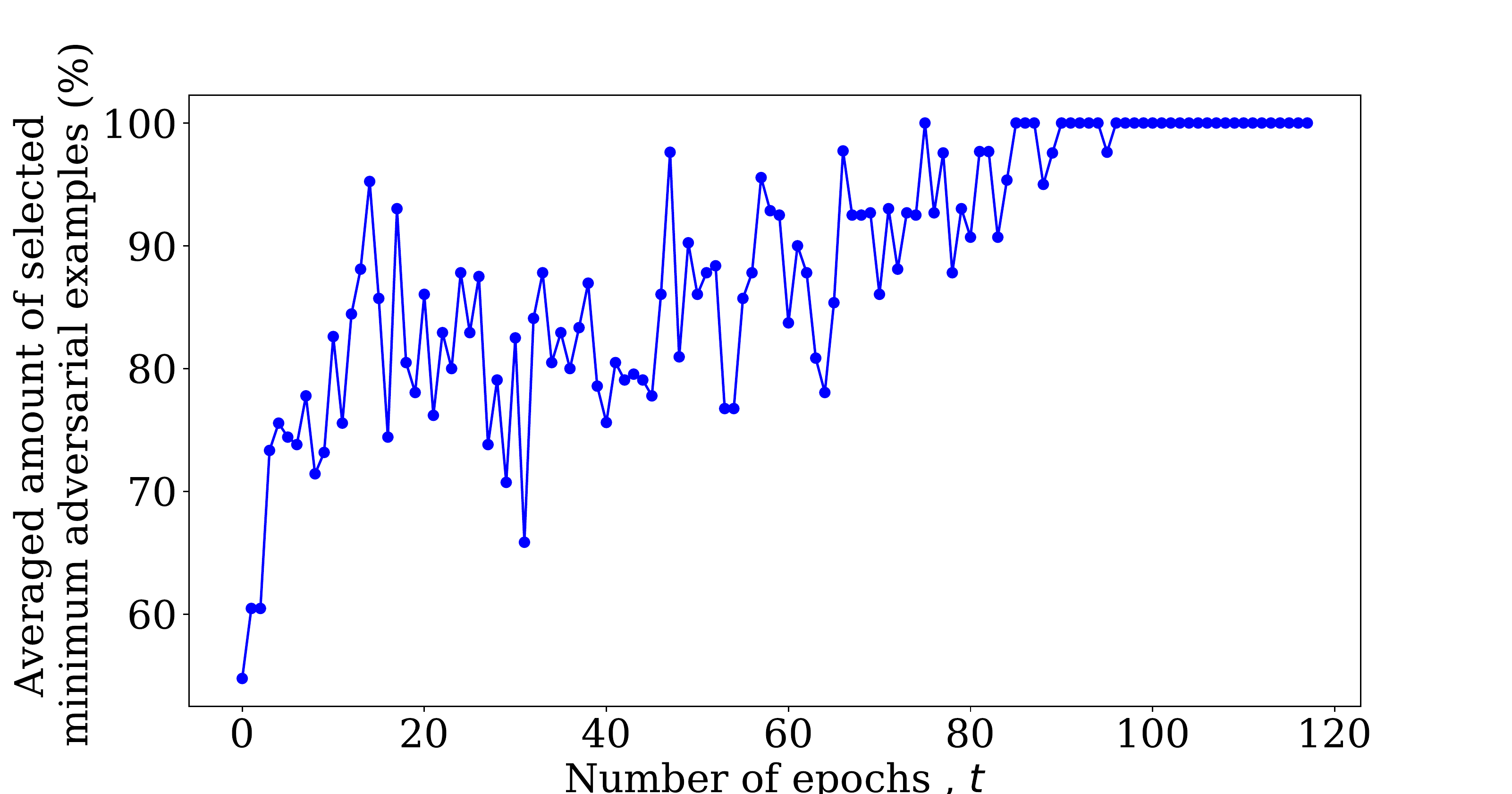}
      \caption{Averaged amount of selected minimum adversarial examples at each epoch for $P_{\rm up} = 0.5$. }
            \label{fig:min_adv}
    \end{figure}

This improvement in robustness-accuracy tradeoff is reasonable
since our method includes the most potential relevant clean and adversarial
samples in the mini-batch. Some
claim that such a tradeoff exists because the standard and robust
objectives conflict \cite{raghunathan2019adversarial,javanmard2020precise}. We can then observe in Figure~\ref{fig:av} that the model trained with $P_{\rm up} = 0.5$ starts by selecting more adversarial samples than clean samples. However, after a few epochs, this behavior changes, and the number of selected clean samples increases. This feature potentially suggests that the model tries to learn the adversarial problem first. When it is done, the DS method attempts to improve the clean accuracy. Moreover, the number of selected minimum adversarial examples increases as the model is trained, as
depicted in Figure~\ref{fig:min_adv}. The minimum adversarial examples are generated by slowly increasing the perturbation constraint $\epsilon$ until the prediction changes.

\begin{figure}[!h]
    \centering
        \includegraphics[width=0.45\textwidth,trim={0 0 0 60},clip]{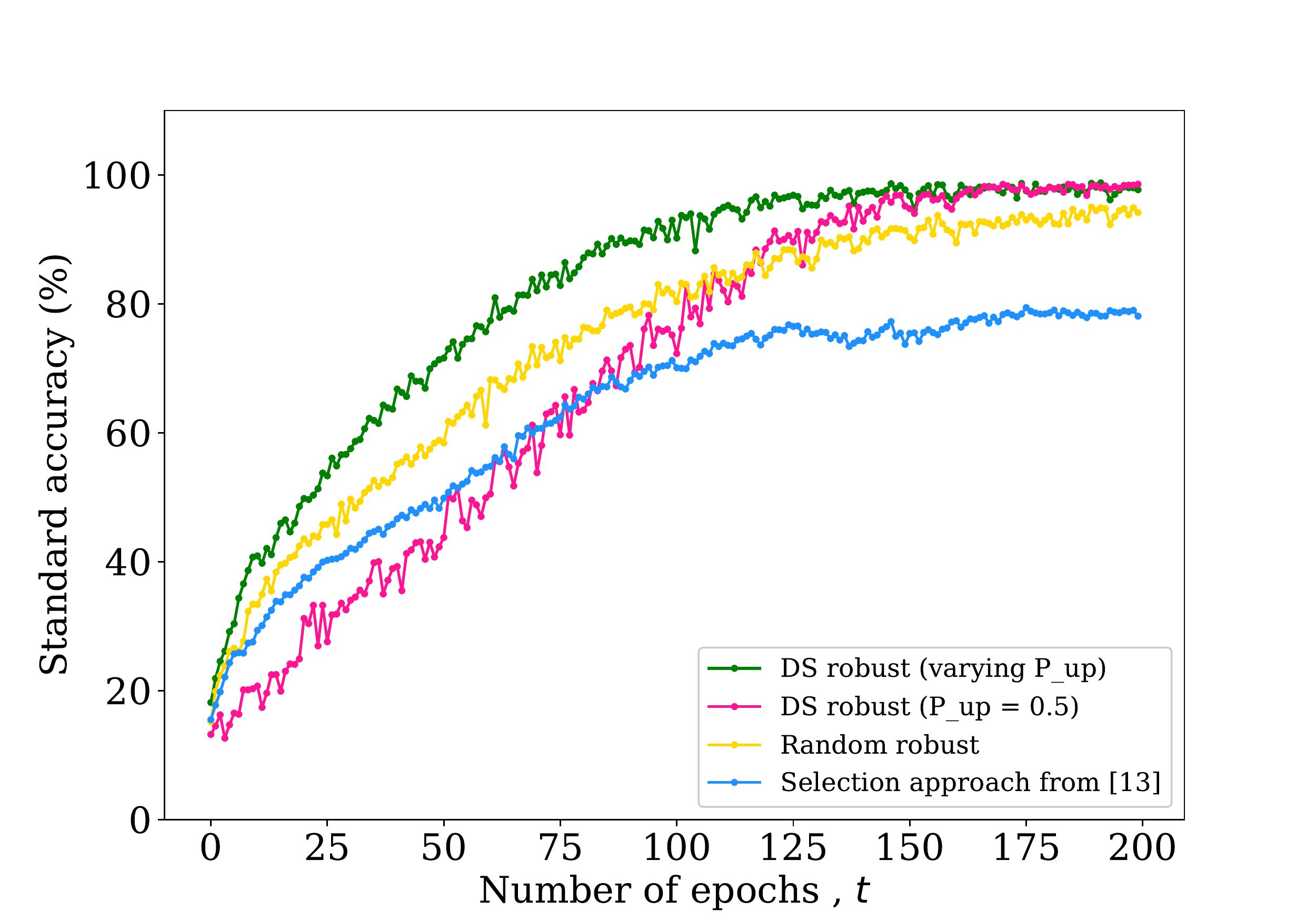}
      \caption{Comparing the proposed method with other selection methods in terms of standard accuracy.  }
            \label{fig:acc_sta_others}
    \end{figure}

\begin{figure}[!h]
    \centering
        \includegraphics[width=0.45\textwidth,trim={0 0 0 60},clip]{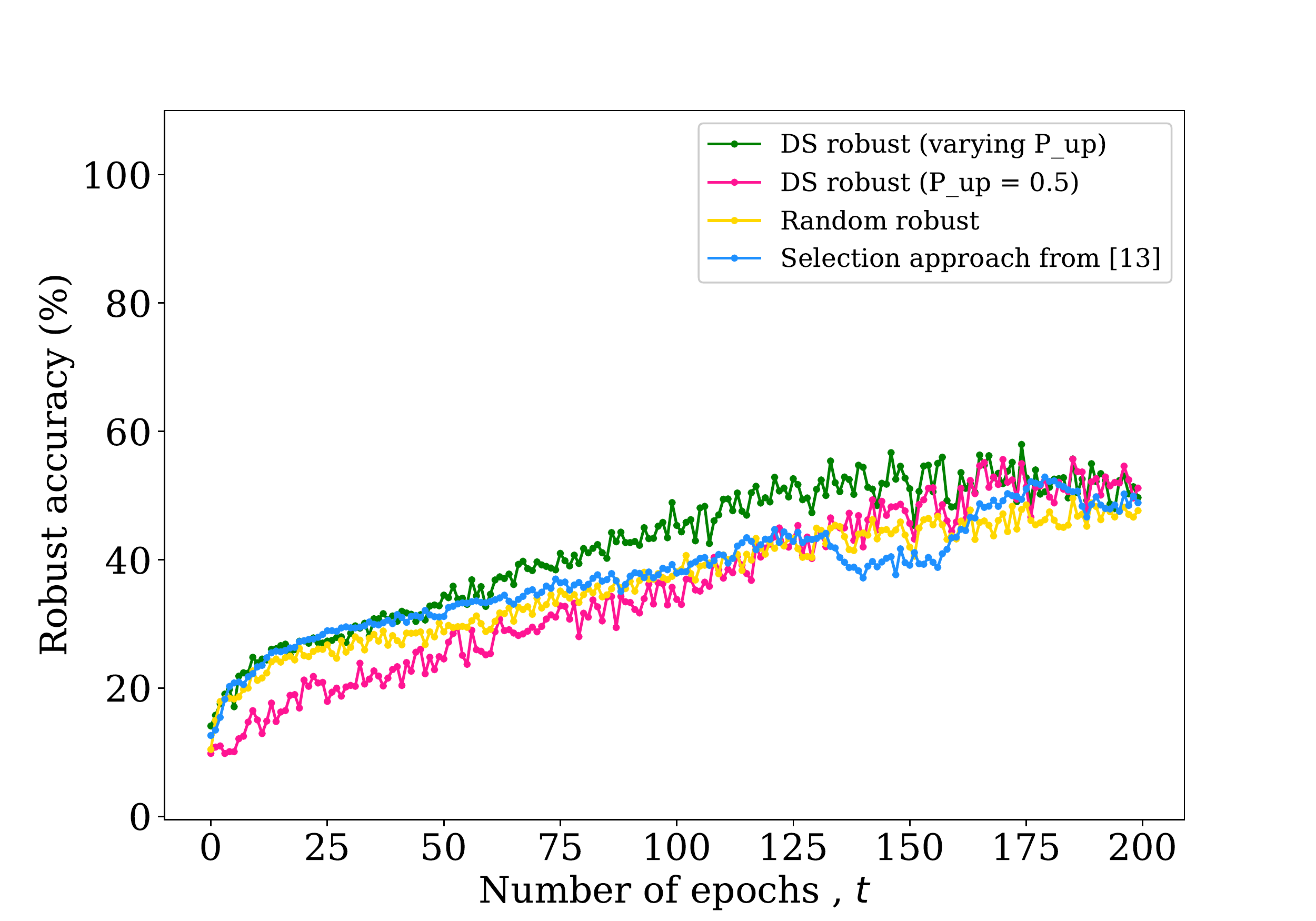}
      \caption{Comparing the proposed method with other selection methods in terms of robust accuracy.  }
            \label{fig:acc_adv_others}
    \end{figure}

Finally, our methods are compared with other selection methods in terms of standard and robust accuracy in Figures \ref{fig:acc_sta_others} and \ref{fig:acc_adv_others}, respectively. The DS approach outperforms both the random method and the selection method with $50 \%$ of high quality samples from \cite{dong2021data}, especially in terms of standard accuracy.

The benefits of the proposed methods in terms of performance are followed by a  reduction in computational complexity. Since only $P_{\rm up}$ samples in the mini-batch are backpropagated through the network to update its parameters; we can save some computations. For example, we present the total training time after 200 epochs in Table~\ref{tab:time}. The simulations were performed in a computer with two GTX-1080 GPUs. With $P_{\rm up} = 0.5$, the training time is reduced when compared with $P_{\rm up} = 1$ and varying $P_{\rm up}$. However, if we stop the training by the 150th epoch, the training time for the varying $P_{\rm up}$ can be reduced to $15261.29$s. Therefore, the varying $P_{\rm up}$ strategy can be applied if an early stopping method is also employed.
We also outperform the method introduced in \cite{dong2021data} in terms of total training time as their method needs a pre-training to rank the samples by the learning stability values.

	{\renewcommand{\arraystretch}{1.4}
\begin{table}[h]
\centering
	\caption{Total training time after 200 epochs.}
			\label{tab:time}
\begin{tabular}{@{}lll@{}}
\toprule \midrule
   Method  &  Time (s)   \\ \midrule
    Selection approach from \cite{dong2021data} with $50 \%$ of samples removed      & 39970.71  \\ \midrule
   Robust with $P_{\rm up} = 1$       & 20200.33  \\ \midrule
   DS Robust with $P_{\rm up} = 0.5$     & 19770.51   \\ \midrule
   DS Robust with $P_{\rm up}$ varying as in equation~(\ref{eq:pup})   & 20161.29   \\ \bottomrule
\end{tabular}
\end{table}}

\section{Conclusion} \label{sec:conc}

Adversarial training is the most popular solution to mitigate the effect of malicious attacks on the deep neural networks. Although adversarial training is able to improve the robustness accuracy, it usually sacrifices standard accuracy in its way. 
Motivated by this drawback and also seeking to reduce the computational complexity during training, we proposed a data selection strategy to include the data samples that bring about a novelty to the learning process.  The simulation results with CIFAR10 using the Resnet18 model indicate that the method is beneficial to improve the robustness-accuracy tradeoff and  reduce the computational complexity of the training.
In the future investigation, one can employ the data selection method to other CNNs models and other datasets.

%

\bibliographystyle{IEEEbib}
\balance
\bibliography{refs}
\end{document}